\documentclass{article}

\PassOptionsToPackage{numbers, compress}{natbib}

\usepackage[preprint]{nips_2018}
\usepackage{color}
\usepackage{tikz}
\usepackage{pgfplots}
\usepackage{pgf-umlsd}
\usepackage{ifthen}
\usepackage{amsmath}
\usepackage{subfiles}
\usepackage{graphicx}
\setcitestyle{square}

\usepackage[utf8]{inputenc} 
\usepackage[T1]{fontenc}    
\usepackage{hyperref}       
\usepackage{url}            
\usepackage{booktabs}       
\usepackage{amsfonts}       
\usepackage{nicefrac}       
\usepackage{microtype}      

\usepackage{natbib}
\title{On Compressing U-net Using Knowledge Distillation }

\author{
  Karttikeya Mangalam\\
  Department of Computer Science\\
  Stanford University\\
  \texttt{mangalam@stanford.edu} \\
   \And
   Dr. Mathieu Salzamann \\
   School of Computer and Communication Science \\
   Ecole Polytechnique Fédérale de Lausanne \\
   \texttt{mathieu.salzmann@epfl.ch} \\
}

\begin{document}

\maketitle

\begin{abstract}
  We study the use of knowledge distillation to compress the U-net architecture. We show that, while standard distillation is not sufficient to reliably train a compressed U-net, introducing other regularization methods, such as batch normalization and class re-weighting, in knowledge distillation significantly improves the training process.
  This allows us to compress a U-net by over 1000x, i.e., to 0.1\% of its original number of parameters, at a negligible decrease in performance.
\end{abstract}

\section{Introduction}
\vspace{-0.2cm}
Deep learning has recently achieved ground-breaking results on several computer vision tasks, yielding highly effective architectures for many common vision tasks, such as image recognition~\cite{vgg} (VGG16, VGG19), human pose estimation~\cite{shg} (Stacked Hourglass Networks) and image segmentation~\cite{unet} (U-net).
However, the increasing number of parameters of these models 
restricts their deployability in resource-constrained environments, such as on mobile devices.
As a consequence, several lines of research have emerged to train more compact networks that achieve a performance similar to the popular, large ones.

In particular, most neural network compression approaches fall in three broad categories:  weight quantization~\cite{quant1,quant2}, architecture pruning~\cite{prun1,prun2,AlvarezSalzmannNIPS17} and knowledge distillation~\cite{caruna,hinton,kd1}. While the first two aim to reduce the size of a given large model by either limiting the number of bits used to represent each parameter, or by removing some of its units or layers, the third, which we investigate here, seeks to train a compact model (student) using knowledge acquired by the larger one (teacher). The first distillation method was introduced by Caruna et al.~\cite{caruna} who focused on the multi-layer perceptron case with an RMSE-based error metric. The concept was then popularized by Hinton et al. \cite{hinton} with an approach exploiting the teacher's predicted probabilities to train the student and by Romero et al.~\cite{kd1} who proposed to further exploit the teacher's intermediate representations to guide the student. Since then, several application-driven knowledge distillation strategies have been developed, e.g., for face model identification~\cite{mobid}, object detection~\cite{objdet} and face verification~\cite{veri}.

In this paper, we study the use of the knowledge distillation technique of~\cite{hinton} to compress a U-net architecture for biomedical image segmentation. We first show that, without performing any distillation, the number of parameters of the U-net can be reduced drastically at virtually no loss of accuracy. We then observe that a direct application of knowledge distillation is insufficient to further compress the U-net, and propose to complement distillation with batch normalization and class re-weighting. As evidenced by our experiments, this allows us to reduce the U-net size to 0.1\% of its original number of parameters at only a negligible loss of segmentation accuracy.

\vspace{-0.15cm}
\section{Methodology}
\begin{figure}[tb]
\centering
\includegraphics[width = \textwidth]{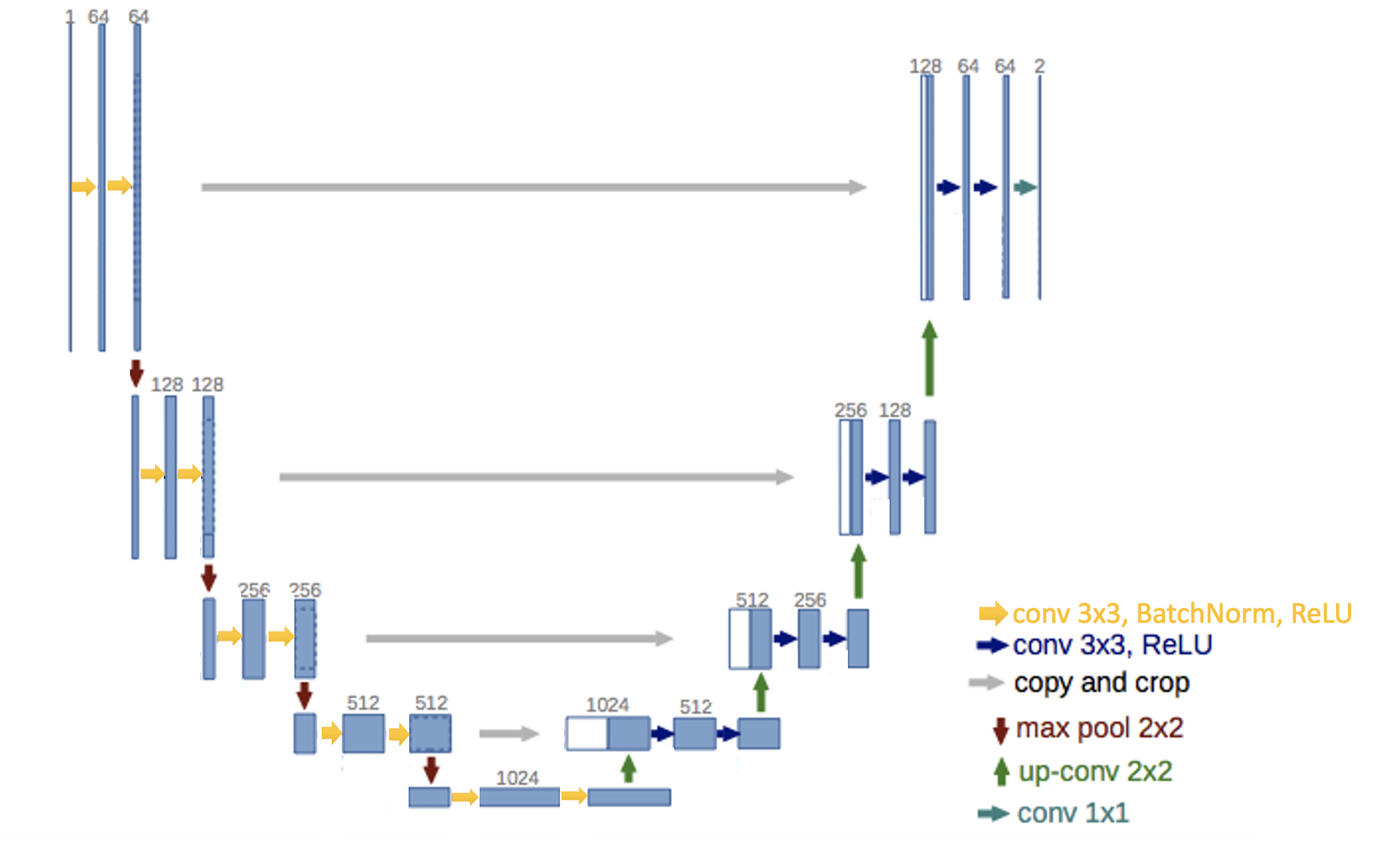}
\caption{U-net architecture. Each blue box corresponds to a multi-channel feature map, with the number of channels indicated on top of the box. 
White boxes represent copied feature maps. Arrows of different colors denote different operations. The golden arrows correspond to our modified version of the original U-net of~\cite{unet}, where we added batch normalization. 
}
\label{fig:unet}
\vspace{-0.0cm}
\end{figure}

\vspace{-0.3cm}
\textbf{U-net Architecture.} The U-net architecture, initially introduced in~\cite{unet} and depicted by Fig.~\ref{fig:unet}, is a fully convolutional network with skip connections, comprising a contracting path and an expansive path. It relies on a channel depth of 64 at the first level and doubles it in 4 consecutive stages, reaching 1024 at the bottom level. This is then reduced back to 64 by the expansive part. 

\begin{table}[t]
\centering
\begin{tabular}{@{}cccccc@{}}
\toprule
Starting Channel Depth & 64     & 16      & 4       & 2              \\ \midrule
Test Loss              & 0.1021 & 0.0822  & 0.0974  & \textbf{0.8227}    \\
Training Loss             & 0.0256 & 0.0220  & 0.0286  & \textbf{0.3423}    \\
Number of Iterations   & 80,000 & 150,000 & 300,000 & 290,000  \\ \bottomrule
\end{tabular}
\vspace{2mm}
\caption{Performance of original U-net architectures (without batch normalization) when varying the channel depth in the first layer. The number of iterations corresponds to the minimum test loss.
}
\label{tab:vanilla}
\vspace{-0.4cm}
\end{table}

\textbf{Knowledge Distillation.}
Let us consider a binary image segmentation problem, where the input is a $W\times H$ image ${\bf x}$ and the output a binary $W\times H$ label map ${\bf y}$. Given training pairs $({\bf x}^t,{\bf y}^t)$, the training procedure of a U-net makes use of the cross-entropy loss
\begin{equation*}
\mathcal{H}(\mathbf{y}^t,\mathbf{\hat{y}}^t) = -\sum_{(i,j)}\mathbf{y}_{ij}^t\log(\mathbf{\hat{y}}_{ij}^t) + (1-\mathbf{y}_{ij}^t)\log(1-\mathbf{\hat{y}}_{ij}^t)\;,
\end{equation*}
where a $\mathbf{y}_{ij}^t$ indicates the label of sample $t$ at location $(i,j)$, and $\mathbf{\hat{y}}^t$ is the probability map predicted by the network. In the standard setting, this probability map is obtained via the softmax function. As discussed in~\cite{hinton}, however, softer probabilities can be obtained by increasing the temperature $T$ of this softmax. Distillation is then achieved by training a teacher network at a temperature $T>1$ to generate probabilities ${\bf \hat{y}}^{t*}$ for each sample ${\bf x}^t$ in a validation set. These probabilities are then employed to train the student network using the cross-entropy $\mathcal{H}(\mathbf{y}^{t*},\mathbf{\hat{y}}^t)$. This cross-entropy can be used either on its own, for vanilla distillation, or in conjunction with the original cross-entropy $\mathcal{H}(\mathbf{y}^t,\mathbf{\hat{y}}^t)$, for mixed distillation. After being trained at $T > 1$, the softmax temperature of the student network is reduced back to $1$.

\textbf{Improving Distillation.}
As evidenced by our results in Section~\ref{sec:exp}, using standard distillation, whether vanilla or mixed, did not prove sufficient to distill a standard U-net into a very small one. To overcome this, we therefore propose two modifications of the original strategy. First, as indicated in Fig.~\ref{fig:unet}, we introduce batch normalization \cite{bn} operations in every convolution layer of the contracting path.
Second, we re-weight the classes according to their proportions in the training set. Specifically, the contribution of each foreground pixel in the loss is multiplied by a weight $w_{\textit{f}}$ equal to the ratio of the number of background pixels over the number of foreground ones. In practice, because there are many more background pixels, this weight is larger than 1, i.e., $w_{\textit{f}}=17.8$. The contribution of the background pixels is kept unchanged, i.e., $w_{\textit{b}}=1.0$.

These two modifications, i.e., batch normalization and class re-weighting, are performed on the student network only, to further help overcoming the general difficulty in training shallow networks. The former reduces the internal covariate shift and the latter combats the inherent class imbalance. 
As shown below, these two modifications allowed us to significantly reduce the necessary U-net size for accurate segmentation.

\vspace{-4mm}
\begin{table}[t]
\centering
\begin{tabular}{@{}ccccc@{}}
\toprule
$T_\text{transfer}$ & Test Loss  & Training Loss (Hard) & Training Loss (Soft) \\ \midrule
2           & 0.490           & 0.268                & 0.167                \\
5           & 0.505            & 0.267                & 0.239                \\
10          & 0.506           & 0.268                & 0.305                \\
15          & 0.507             & 0.270                & 0.326                \\
20          & 0.507          & 0.271                & 0.334                \\ \bottomrule
\end{tabular}
\vspace{2mm}
\caption{Performance of the 2-Unet using mixed distillation with class re-weighting (without batch normalization). 
The training soft losses are reported after multiplying by $T^2_\textit{transfer}$ to adjust for scale.  }
\label{tab:soft-hard}
\vspace{-0.2cm}
\end{table}

\begin{table}[]
\centering
\begin{tabular}{@{}cccc@{}}
\toprule
Network                     & \# trainable parameters & IoU  Score & Cross Entropy Loss \\ \midrule
Original 64-Unet                     & 31,042,434
& 0.804                         & 0.102   \\
Original 4-Unet                     & 122,394
&             \textbf{0.807}             & \textbf{0.097}
\\
Our 2-Unet  (soft loss only)    & \textbf{30,902}        & 0.752                       & \underline{0.134}              \\
Our 2-Unet (mixed distillation) & 30,902        & \underline{0.759}                         & 0.135              \\ \bottomrule
\end{tabular}
\vspace{2mm}\caption{Comparison of our method with a 64-Unet. Note that we achieve a competitive Intersection over Union Score to the original 64-Unet with only \textbf{0.1\%} of its capacity. }
\label{tab:final_results}
\vspace{-0.4cm}
\end{table}

\vspace{0.2cm}
\section{Experiments}
\label{sec:exp}
\vspace{-0.2cm}

\textbf{Dataset.} 
We use the Electron Microscopy (EM) Mitochondria Segmentation dataset of~\cite{ds1}. It contains a 5x5x5 micrometer section taken from the CA1 hippocampus region of the brain, corresponding to 1065x2048x1536 voxels of resolution approximately 5x5x5nm. This volume is separated into training and testing sub-volumes, each of which consists of 165 slices. We treat each slice as an image and aim to produce the corresponding binary label map, indicating the presence or absence of a mitochondria at the each location.

\textbf{Compressing without Distillation.} 
First, we experiment with reducing the number of channels in the first U-Net layer while keeping the doubling trend of the contracting path without any distillation. As shown in Table~\ref{tab:vanilla}, a U-net with only 4 initial channels (4-Unet) achieves a similar test loss to the original 64-Unet. In the remaining experiments, we use the 4-Unet as teacher network for distillation.

\textbf{Exploiting Standard Distillation.} We then tried to make use of the standard distillation procedure of~\cite{hinton} to train a 2-Unet from a 4-Unet. To this end, we evaluated both vanilla distillation, mixed distillation and sequential distillation, where we started training with soft labels and then finished with hard ones, or vice-versa. All these attempts were unsuccessful, even with class re-weighting. This is illustrated by Table~\ref{tab:soft-hard} for mixed distillation at different temperatures.

\textbf{Distilling Our Modified U-net.} 
Finally, we evaluate the use of distillation with our modified U-net that incorporates batch normalization and class re-weighting. Note that for all the distillation experiments, the teacher is the original 4-Unet trained from scratch with hard training loss.

As shown in Table~\ref{tab:final_results}
for both vanilla distillation and mixed distillation, a 2-Unet achieves segmentation accuracies similar to those of a standard 64-Unet, while requiring only about 0.1\% of its capacity. Note that training a 2-Unet without distillation but with our modifications yields a training loss of 0.265 and a test loss of 0.307, which shows that distillation is still required.

\vspace{-0.2cm}
\section{Conclusion}
\vspace{-0.2cm}
We have introduced a modified distillation strategy to compress a U-net architecture by over 1000x while retaining an accuracy close to the original U-net. This was achieved by modifying the U-net to incorporate batch normalization and class re-weighting. 
In the future, we plan to investigate the use of these modifications to perform distillation of other networks and for other application domains.

\bibliography{refs}

\end{document}